\documentclass[11pt]{article}

\usepackage[preprint]{acl}
\usepackage{times}
\usepackage{latexsym}
\usepackage[T1]{fontenc}
\usepackage[utf8]{inputenc}
\usepackage{microtype}
\usepackage{inconsolata}
\usepackage{graphicx}
\usepackage{amsmath}
\usepackage{amssymb}
\usepackage{booktabs}
\usepackage{array}
\usepackage{colortbl}
\usepackage[most]{tcolorbox}
\usepackage{algorithm}
\usepackage{algpseudocode}
\usepackage{tikz}
\usepackage{pgfplots}
\usetikzlibrary{arrows.meta,positioning,fit,shapes.geometric}
\usepgfplotslibrary{groupplots}
\pgfplotsset{compat=1.18}

\newcommand{\method}{\textsc{ReSkill}}
\newcommand{\ind}{\mathbb{I}}
\newcommand{\gain}[1]{\textcolor{gray}{\scriptsize(#1)}}
\definecolor{directrow}{RGB}{238,247,255}
\definecolor{hyprow}{RGB}{255,246,232}
\definecolor{actrow}{RGB}{238,249,240}

\title{\method{}: Explicit Failure Attribution and Structured Repair for Interactive Language Agents}

\author{
Mengyi Deng$^{1}$, Xin Li$^{1}$, Duyi Pan$^{1}$, Zilin Wang$^{1}$, Zhiwei Li$^{1}$,
Zhijiang Guo$^{1,2,\dagger}$, Wei Wang$^{1,2,\dagger}$\\
$^{1}$Information Hub, The Hong Kong University of Science and Technology (Guangzhou),
China \\
$^{2}$The Hong Kong University of Science and Technology, Hong Kong SAR\\
\texttt{\{mdeng974, xli420, dpan457, zwang374, zli404\}@connect.hkust-gz.edu.cn} \\
\texttt{zhijiangguo@hkust-gz.edu.cn, weiwcs@ust.hk}
}

\begin{document}
\maketitle

\begin{abstract}
Language agents increasingly rely on reusable skills, but post-failure repair is often handled by opaque one-shot reflection: a model generates a skill patch without explicitly maintaining how failure explanations relate to candidate repairs or how unsuccessful retests should influence later edits. We introduce \method{}, a structured repair framework that maintains an explicit repair state across repair rounds. Given a failed rollout, the framework links failure hypotheses to candidate skill patches, selects local repairs through coverage-based attribution, retests the edited skill set in the environment, and uses retest outcomes to guide subsequent repair updates. The language model supplies structured repair factors, while the repair procedure records them, compares local skill patches by how well they address active failure explanations, and carries unsuccessful retest outcomes into later repair rounds. We evaluate \method{} on ALFWorld and TextCraft across three model sizes under fixed repair budgets. \method{} obtains the strongest final success in all six benchmark--model settings, improving average final success by 3.7 percentage points over direct repair and 3.3 points over hypothesis-conditioned repair. These results suggest that explicit attribution alone is insufficient; durable improvement emerges when attribution is integrated with repair selection and persistent retest-conditioned update. 

\end{abstract}

\section{Introduction}
\label{sec:introduction}

Interactive language agents increasingly rely on reusable skills and behavioral rules for long-horizon tool-use and embodied interaction~\cite{wang2023voyager,yang2026prune}. Prior work has studied reusable skill acquisition~\cite{wang2023voyager,chen2026skillcraft}, iterative self-improvement, and continual skill evolution~\cite{liu2026skillforge} from reflection and interaction experience~\cite{zheng2504skillweaver,yang2026autoskill}. Despite this progress, reusable skills remain brittle, and repairing them after failure is difficult because a single failed rollout may point to multiple plausible, potentially co-occurring causes, such as a missing precondition, incorrect object handling, or an invalid action. Because candidate edits may address different subsets of these causes, repair requires deciding which edit to apply and, after an unsuccessful retest, which explanations should guide the next repair round.

However, existing repair approaches often handle post-failure correction by asking a language model to rewrite the skill set directly from a failed trajectory~\citep{shinn2023reflexion,madaan2023selfrefine,liu2026skillforge,patel2024large}. While such approaches can produce useful local patches and progressively improve reusable skills~\cite{shen2026skillfoundry,zhang2026coevoskills,yang2026autoskill}, they typically conflate diagnosis, repair generation, and repair selection into a single generation step. As a result, post-failure repair becomes a repair credit-assignment problem: a failed trajectory may support several plausible explanations, but an effective edit may address only a subset of them. If the system does not record which explanations each patch is meant to address, or how retest feedback changes support for those explanations, iterative repair can repeatedly apply ineffective edits or overfit to a spurious diagnosis. Effective repair therefore requires reducing failure ambiguity, attributing observed outcomes to the explanations under repair, and carrying credit across repair rounds.

We formulate post-failure skill repair as hypothesis-conditioned repair selection under environment feedback. Given a failed trajectory, a language model generates a support set of failure hypotheses, candidate skill patches, and a coverage matrix describing which patches address which hypotheses. The repair algorithm tracks prioritized failure hypotheses, compares local patches by how well they cover the active hypotheses, applies the selected patch to the skill set, and retests the task in the environment. If a retest fails, the resulting post-repair trace is converted into updated hypotheses for the next repair round. Figure~\ref{fig:framework}(b) illustrates these components in a concrete ALFWorld example. The agent is asked to cool a mug and place it on the coffee machine, but instead cools an egg and repeatedly examines the mug without picking it up. From this evidence, \method{} forms two active hypotheses: incorrect object selection and a missing pickup action. It selects a patch that requires picking up the mug before cooling it, addressing both hypotheses, and validates the repair through a successful retest.

We evaluate \method{} in two interactive settings that stress different forms of skill use: grounded household interaction in ALFWorld~\citep{shridhar2020alfworld} and symbolic crafting in TextCraft from AgentGym~\citep{agentgym2024}. Across the resulting six benchmark--model settings, \method{} improves final repair success under fixed repair budgets, with average gains of 3.7 and 3.3 percentage points over two competitive repair baselines. The results suggest that grounding repair decisions in hypothesized failure causes, comparing candidate edits against competing explanations, and revising repair credit through environment retesting improve post-failure skill repair across model scales. In summary, our contributions are:
\begin{itemize}
    \item We formulate post-failure skill repair as a repair credit-assignment problem over ambiguous failure explanations, candidate edits, and retest outcomes.
    \item We introduce \method{}, a structured framework that links failure hypotheses, candidate repairs, coverage-based selection, environment retesting, and outcome-conditioned repair updates.
    \item We instantiate \method{} in ALFWorld and TextCraft without changing model weights, showing improvements across three model sizes and providing implementation-grounded analyses of repair traces.
\end{itemize}

\section{Related Work}
\label{sec:related}
\paragraph{Reusable skill for agents.}
Recent language-agent systems increasingly externalize reusable behavior into skills, tool calls, or programmatic action interfaces. In robotics and tool use, language models can be grounded through affordances, APIs, or code-like policies~\citep{ahn2022can,huang2022inner,liang2023code,huang2024effilearner,xu2026envfactory,deng2026uncertainty,zhang2025system}. In open-ended embodied environments, agents accumulate competence through executable skill libraries, retrieved memories, or structured action knowledge~\citep{wang2023voyager,zhu2023ghost,wang2024jarvis,liu2024odyssey,fan2022minedojo,park2023generative,liu2026well}. 

Recent skill-centric work makes this substrate more explicit: SkillAct studies prompting agents with reusable skill abstractions~\citep{liu2024skillact}; SKILL0 studies internalizing skill context through an in-context reinforcement-learning curriculum~\citep{lu2026skill0}; SkillX constructs transferable skill knowledge bases from trajectories~\citep{wang2026skillx}; and SkillGen and SkillOS study skill synthesis and curation from execution experience~\citep{ma2026skillgen,ouyang2026skillos}. These works establish reusable skills as a practical interface for long-horizon behavior. \method{} builds on this view by studying the post-failure repair stage: how failed interactions can be attributed to competing skill-level explanations and converted into targeted skill updates.

\paragraph{Feedback-driven self-improvement.}
Reflection methods show that feedback can improve future behavior. ReAct interleaves reasoning and acting during interaction~\citep{yao2022react}; DEPS uses language-model explanation and selection for open-world planning~\citep{wang2023describe}; Reflexion stores verbal reflections after failed trials~\citep{shinn2024reflexion}; and Self-Refine iteratively improves outputs using self-feedback~\citep{madaan2023self}. 

Related work also separates generation from selection. Tree of Thoughts expands multiple reasoning paths before choosing among them~\citep{yao2023tree}; SayCan combines language scores with affordance values for action selection~\citep{ahn2022saycan}; and tool-use methods select or invoke external capabilities as part of solving a task~\citep{schick2023toolformer}. These methods demonstrate the value of separating proposal generation from downstream selection. However, they do not extend this separation to post-failure skill repair, where candidate edits must be linked to competing failure hypotheses. \method{} makes this link explicit through hypothesis--patch coverage and environment retesting.

\section{Method}
\label{sec:method}

\begin{figure*}[!t]
    \centering
    \includegraphics[width=1\linewidth]{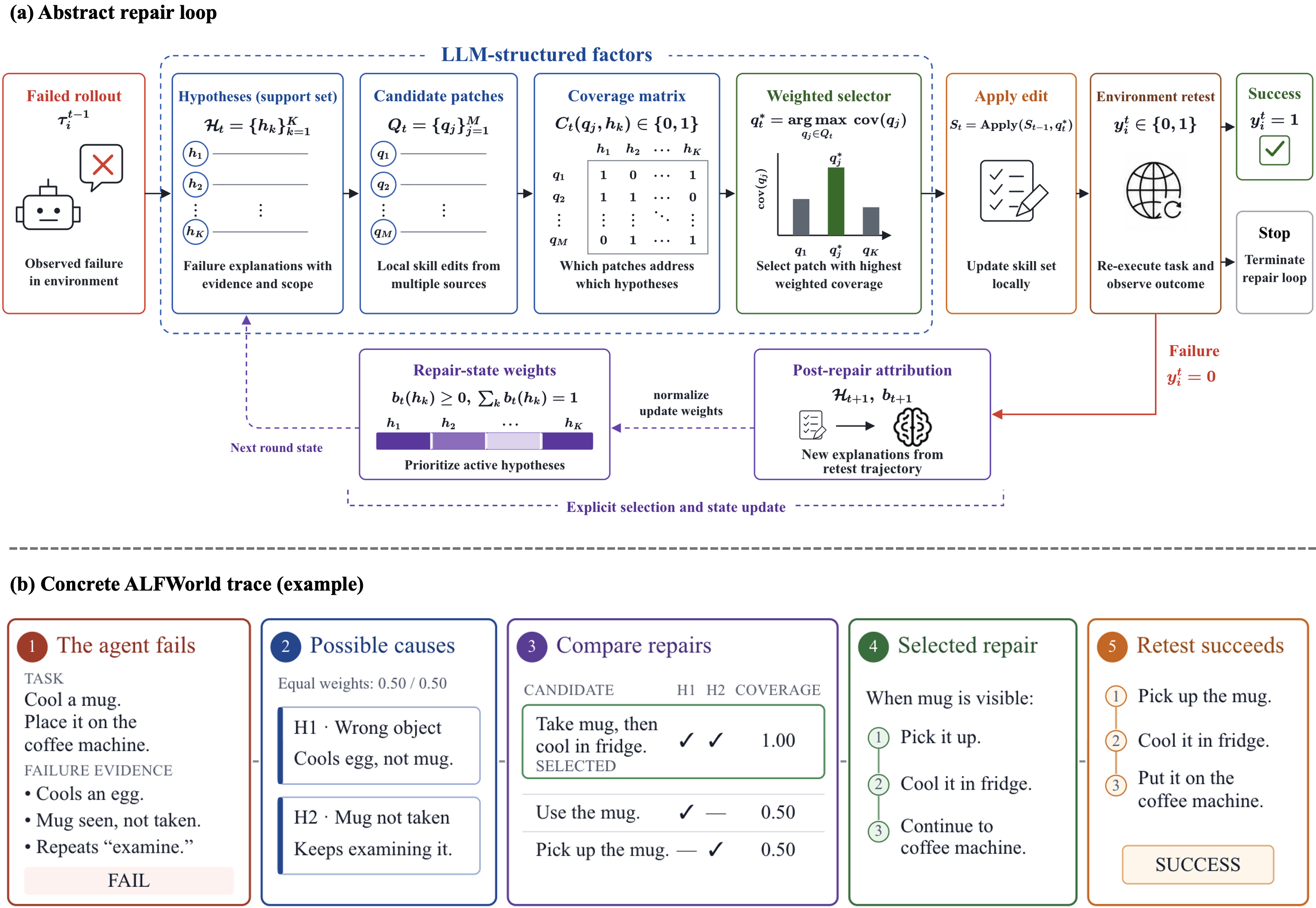}
    \caption{
Overview of \method{} for structured post-failure skill repair.
(a) Given a failed rollout, the framework constructs a set of failure hypotheses, generates candidate skill patches, selects a patch by weighted hypothesis coverage, and updates the hypothesis state through environment retesting.
(b) A concrete ALFWorld example showing how failure evidence gives rise to multiple hypotheses and candidate patches. The repair that best addresses the active hypotheses is then selected and validated through environment retesting.
}
    \label{fig:framework}
\end{figure*}
\method{} is a post-failure repair framework for editable agent skills. As shown in Figure~\ref{fig:framework}, given a failed trajectory, the language model proposes multiple failure hypotheses and local candidate patches, and identifies which hypotheses each patch addresses. \method{} maintains weights over the active hypotheses, selects a patch based on its weighted coverage, and retests the edited skill set in the environment. If the retest succeeds, repair stops; otherwise, the failed retest trajectory is used to update the hypotheses and their weights for the next round.

\subsection{Post-Failure Skill Repair}

Let \(x_i\) denote an interactive task, \(S_0\) an initial skill set, and \(\pi_\theta(\cdot\mid S)\) the frozen language-agent policy induced by prompting the model with skill set \(S\). Model parameters \(\theta\) remain fixed throughout repair; repair modifies only the external skill set used to condition the policy. The same underlying frozen language model is used for task execution and structured repair generation through separate prompts and calls. The first pass executes a complete rollout under \(S_0\), producing an interaction trajectory
\[
\tau_i^0=(o_0,a_0,o_1,\ldots,a_{K_i^0-1},o_{K_i^0})
\]
and a binary success indicator \(y_i^0\in\{0,1\}\). Here \(o_k\) is the environment observation after \(k\) actions, \(a_k\) is the action emitted after observing \(o_k\), and \(K_i^0\) is the number of agent actions before the episode terminates or reaches the task budget. Repair is triggered only after a completed unsuccessful rollout, not during intermediate action steps.

At repair round \(t\ge 1\), provided that \(y_i^{t-1}=0\), \method{} uses the failed trajectory \(\tau_i^{t-1}\) as evidence to select a patch \(q_t\). Applying the patch changes the skill set locally from \(S_{t-1}\) to:
\[
S_t = \mathrm{Apply}(S_{t-1}, q_t).
\]
The operator \(\mathrm{Apply}\) inserts the patch into the skill section identified by its target scope and leaves unrelated sections unchanged. The agent then retests the task under \(S_t\), producing a new trajectory \(\tau_i^t\) and success indicator \(y_i^t\).
The repair budget \(R\) limits the number of post-failure retests. For an evaluation set of \(N\) tasks and any budget \(r\in\{0,\ldots,R\}\), we report cumulative task success, including both first-pass successes and tasks recovered within \(r\) repair rounds:
\[
\mathrm{Succ}@r =
\frac{1}{N}\sum_{i=1}^N
\ind\left[y_i^0=1 \vee \exists t\le r:\; y_i^t=1\right].
\]
Tasks solved on the first pass count at every budget, while initially failed tasks can be recovered by any later repair retest. Each attempted patch is evaluated by re-executing the triggering task under the edited skill set. The selected patch, retest trajectory, and outcome are recorded for the next repair round.

\subsection{Structured Repair State}

At repair round \(t\), \method{} organizes the information needed for the next repair decision into a structured state rather than directly rewriting the skill set from the failed trajectory. This state contains the current skill set \(S_{t-1}\), the latest failed trajectory \(\tau_i^{t-1}\), the active hypothesis set \(\mathcal{H}_t\) with weights \(b_t(h)\), and the history of previously attempted patches and their retest outcomes. Using this state, the language model generates candidate local patches and coverage relations indicating which active hypotheses each patch addresses. \method{} selects the patch \(q_t^\star\) with the highest weighted coverage, applies it to the skill set, and retests the task in the environment. If the retest fails, the new trajectory is used to update the hypothesis set and its weights, while the attempted patch and its outcome are recorded for the next repair round.

\subsection{Hypothesis State}

\method{} represents a failed trajectory as a finite support of possible failure explanations. At the first repair round, the model produces an initial hypothesis set \(\mathcal{H}_1\). After an unsuccessful retest, post-repair attribution produces the next set \(\mathcal{H}_{t+1}\). At round \(t\), the active support is
\[
\mathcal{H}_t = \{h_1,\ldots,h_K\},
\]
where each hypothesis contains a concise failure explanation, supporting
evidence from the trajectory, the affected skill scope, and a repair hint.
Across rounds, hypotheses are aligned when they represent the same underlying
failure cause within the same skill scope; hypotheses that cannot be aligned
with an earlier explanation are treated as newly introduced. \method{}
maintains a normalized weight \(b_t(h)\) for each \(h\in\mathcal{H}_t\). The first repair round starts from uniform weights,
\[
b_1(h)=\frac{1}{|\mathcal{H}_1|},
\qquad h\in\mathcal{H}_1,
\]
and later rounds update the weights from the retest outcome. These weights determine which active explanations receive priority when the selector compares candidate edits. A high-weight hypothesis contributes to every candidate that covers it, allowing the selector to compare narrower and broader local edits under the same coverage-based comparison.

\subsection{Repair Candidates}

Given the latest failed trajectory \(\tau_i^{t-1}\), the current skill set \(S_{t-1}\), the active hypotheses \(\mathcal{H}_t\), and the patches attempted in earlier rounds, \method{} builds a finite candidate pool
\(\mathcal{Q}_t\):
\[
\mathcal{Q}_t =
\mathcal{Q}_t^{\mathrm{trace}}
\cup
\mathcal{Q}_t^{\mathrm{hyp}}
\cup
\mathcal{Q}_t^{\mathrm{prop}}
\cup
\mathcal{Q}_t^{\mathrm{guard}} .
\]
The superscripts index candidate sources used within a single \method{} round. \(\mathcal{Q}_t^{\mathrm{trace}}\) contains a trajectory anchor patch written from the observed failure evidence. \(\mathcal{Q}_t^{\mathrm{hyp}}\) contains a hypothesis-conditioned anchor patch written with access to the active support \(\mathcal{H}_t\). \(\mathcal{Q}_t^{\mathrm{prop}}\) contains additional targeted proposals that cover different parts of \(\mathcal{H}_t\), including edits that intentionally address overlapping hypotheses. \(\mathcal{Q}_t^{\mathrm{guard}}\) contains interface-preserving guard candidates used when a local grounding, admissibility, or executable-format correction can be derived from the observed trace.

Each candidate \(q\in\mathcal{Q}_t\) is represented as a local skill patch with an intended scope, the evidence it claims to address, and the skill text to be inserted or revised. Candidates are screened before selection: the patch must be reusable beyond the current episode, grounded in the observed trajectory, compatible with the agent's action interface, and narrow enough to avoid rewriting unrelated skills. The pool therefore gives the selector alternatives to compare while keeping repair local and bounded.

\subsection{Coverage-Based Selection}

For each candidate patch \(q\in\mathcal{Q}_t\) and each hypothesis \(h\in\mathcal{H}_t\), the model assigns a binary coverage indicator,
\[
C_t(q,h)\in\{0,1\}.
\]
The model assigns \(C_t(q,h)=1\) only when the proposed edit directly addresses the failure cause represented by \(h\) and applies to the relevant skill scope. Before selection, candidate patches are required to specify a local, executable edit grounded in the failed trajectory. The selector scores each candidate by the active hypothesis weight it covers:
\[
\mathrm{cov}_t(q)
=
\sum_{h\in\mathcal{H}_t} b_t(h) C_t(q,h).
\]
A patch can therefore receive a high score by addressing either one high-weight hypothesis or several lower-weight hypotheses. The selected patch is:
\[
q_t^\star
=
\arg\max_{q\in\mathcal{Q}_t}\mathrm{cov}_t(q),
\]
with ties resolved in favor of more local and evidence-grounded edits. This design separates patch generation from the final repair decision: the model proposes candidate edits and their coverage relations, while \method{} applies an explicit coverage-based selection rule.

\subsection{Outcome-Conditioned Update}

After applying \(q_t^\star\), the agent retests the task in the environment. If \(y_i^t=1\), repair terminates. Otherwise, the model uses the failed retest trajectory \(\tau_i^t\), the previous hypotheses, and the attempted patch to construct \(\mathcal{H}_{t+1}\) for the next repair round. If no post-repair hypothesis supported by the retest trajectory is obtained, \method{} retains \(\mathcal{H}_t\) and \(b_t\) while continuing from the latest failed trajectory. \method{} aligns recurring explanations with their counterparts from the previous round and treats unmatched explanations as newly introduced. The total active hypothesis weight covered by the selected patch is
\[
m_t
=
\sum_{h\in\mathcal{H}_t} b_t(h)C_t(q_t^\star,h).
\]
This quantity determines the initial weight assigned to explanations newly exposed by an unsuccessful retest. Intuitively, when a patch addressing active hypotheses still fails, the retest increases the relative priority of explanations newly exposed after that intervention. Recurring hypotheses retain their previous weights, while \(m_t\) is divided evenly among newly introduced hypotheses. The resulting weights are then normalized over \(\mathcal{H}_{t+1}\). If all assigned weights are zero, \method{} uses uniform weights. The next candidate pool \(\mathcal{Q}_{t+1}\) is generated from the failed retest trajectory, the updated hypotheses, and the patches attempted in earlier rounds.

\begin{algorithm}[t]
\caption{\method{} for Post-Failure Skill Repair}
\label{alg:act-repair}
\small
\begin{algorithmic}[1]
    \State \textbf{Input:} task \(x_i\), initial skill set \(S_0\)
    \Statex \hspace{\algorithmicindent} repair budget \(R\), frozen policy \(\pi_\theta\)
    \State \textbf{Output:} final skill set, success flag,
    \Statex \hspace{\algorithmicindent} and repair trace
    \State Roll out \(\pi_\theta(\cdot\mid S_0)\) to obtain \(\tau_i^0\) and \(y_i^0\)
    \If{\(y_i^0=1\)}
        \State \Return \(S_0\), success, first-pass trace
    \EndIf
    \State Generate initial hypotheses \(\mathcal{H}_1\) from \(\tau_i^0\)
    \State Initialize \(b_1(h)=1/|\mathcal{H}_1|\) for \(h\in\mathcal{H}_1\)
    \For{\(t=1,\ldots,R\)}
        \State Build candidate repairs \(\mathcal{Q}_t\) from the current trace and repair state
        \State Screen candidates for locality, grounding, validity,
        \Statex \hspace{\algorithmicindent} and non-redundancy
        \State Construct repair--hypothesis coverage relations \(C_t(q,h)\)
        \Statex \hspace{\algorithmicindent} for \(q\in\mathcal{Q}_t\) and \(h\in\mathcal{H}_t\)
        \State Select \(q_t^\star\) using covered weight and fixed locality/grounding criteria
        \State Apply the patch: \(S_t\gets\mathrm{Apply}(S_{t-1},q_t^\star)\)
        \State Retest with \(S_t\) to obtain \(\tau_i^t\) and \(y_i^t\)
        \If{\(y_i^t=1\)}
            \State \Return \(S_t\), success, repair trace
        \EndIf
        \State Generate post-repair hypotheses \(\mathcal{H}_{t+1}\)
        \If{\(\mathcal{H}_{t+1}\) contains usable hypotheses}
            \State Update \(b_{t+1}\) over \(\mathcal{H}_{t+1}\) using the outcome-conditioned update
        \Else
            \State Carry forward the previous repair state
        \EndIf
    \EndFor
    \State \Return \(S_R\), failure, repair trace
\end{algorithmic}
\end{algorithm}

Algorithm~\ref{alg:act-repair} summarizes the repair loop for one task. Each attempted repair records the active hypotheses, candidate patches, coverage relations, selected patch, retest outcome, and next-round state. When a retest fails, these records guide the next repair decision by preserving
persistent explanations and incorporating newly exposed ones.

\section{Experiments}
\label{sec:experiments}

\subsection{Experimental Setting}

\begin{table*}[t]
\centering
\small
\setlength{\tabcolsep}{1.2pt}
\renewcommand{\arraystretch}{1.07}
\resizebox{\textwidth}{!}{
\begin{tabular}{llrrrrrrrrrr}
\toprule
& & \multicolumn{6}{c}{\textbf{ALFWorld}} & \multicolumn{3}{c}{\textbf{TextCraft}} \\
\cmidrule(lr){3-8}\cmidrule(lr){9-11}
Model & Method & Overall & Pick & Look & Clean & Heat & Cool & Overall & D1 & D2 \\
\midrule
Qwen3-1.7B & No skills & 7.7 & 9.0 & 19.4 & 5.2 & 2.6 & 4.3 & 26.3 & 70.9 & 5.6 \\
Qwen3-1.7B & With skills & 16.8 & 9.0 & 35.5 & 27.6 & 12.8 & 10.9 & 21.1 & 74.6 & 5.2 \\
\rowcolor{directrow}
Qwen3-1.7B & Direct & 30.7 \gain{+13.9} & 22.0 \gain{+13.0} & 48.4 \gain{+12.9} & 46.6 \gain{+19.0} & \textbf{30.8} \gain{+18.0} & \textbf{17.4} \gain{+6.5} & 28.5 \gain{+7.4} & 85.1 \gain{+10.5} & 13.9 \gain{+8.7} \\
\rowcolor{hyprow}
Qwen3-1.7B & Hypothesis & 27.7 \gain{+10.9} & 18.0 \gain{+9.0} & 48.4 \gain{+12.9} & 48.3 \gain{+20.7} & 23.1 \gain{+10.3} & 13.0 \gain{+2.1} & 28.9 \gain{+7.8} & 86.6 \gain{+12.0} & 14.2 \gain{+9.0} \\
\rowcolor{actrow}
Qwen3-1.7B & \method{} & \textbf{33.2} \gain{+16.4} & \textbf{25.0} \gain{+16.0} & \textbf{58.1} \gain{+22.6} & \textbf{53.4} \gain{+25.8} & 23.1 \gain{+10.3} & \textbf{17.4} \gain{+6.5} & \textbf{30.7} \gain{+9.6} & \textbf{91.8} \gain{+17.2} & \textbf{15.3} \gain{+10.1} \\
\midrule
Qwen2.5-3B & No skills & 9.9 & 9.0 & 25.8 & 6.9 & 5.1 & 8.7 & 13.0 & 29.9 & 5.2 \\
Qwen2.5-3B & With skills & 22.6 & 22.0 & 45.2 & 17.2 & 17.9 & 19.6 & 9.9 & 29.9 & 4.9 \\
\rowcolor{directrow}
Qwen2.5-3B & Direct & 53.3 \gain{+30.7} & \textbf{50.0} \gain{+28.0} & 74.2 \gain{+29.0} & \textbf{67.2} \gain{+50.0} & 41.0 \gain{+23.1} & 39.1 \gain{+19.5} & 17.5 \gain{+7.6} & 50.7 \gain{+20.8} & 9.4 \gain{+4.5} \\
\rowcolor{hyprow}
Qwen2.5-3B & Hypothesis & 52.2 \gain{+29.6} & 47.0 \gain{+25.0} & \textbf{77.4} \gain{+32.2} & 51.7 \gain{+34.5} & 51.3 \gain{+33.4} & \textbf{47.8} \gain{+28.2} & 22.8 \gain{+12.9} & 61.2 \gain{+31.3} & 14.2 \gain{+9.3} \\
\rowcolor{actrow}
Qwen2.5-3B & \method{} & \textbf{54.7} \gain{+32.1} & 47.0 \gain{+25.0} & \textbf{77.4} \gain{+32.2} & 58.6 \gain{+41.4} & \textbf{59.0} \gain{+41.1} & \textbf{47.8} \gain{+28.2} & \textbf{29.4} \gain{+19.5} & \textbf{85.8} \gain{+55.9} & \textbf{15.6} \gain{+10.8} \\
\midrule
Qwen3-4B & No skills & 54.7 & 73.0 & 48.4 & 37.9 & 64.1 & 32.6 & 42.4 & 81.3 & 24.3 \\
Qwen3-4B & With skills & 80.7 & 82.0 & 93.5 & 89.7 & 61.5 & 73.9 & 47.4 & 87.3 & 28.8 \\
\rowcolor{directrow}
Qwen3-4B & Direct & 93.8 \gain{+13.1} & 96.0 \gain{+14.0} & \textbf{100.0} \gain{+6.5} & 96.6 \gain{+6.9} & \textbf{92.3} \gain{+30.8} & 82.6 \gain{+8.7} & 48.1 \gain{+0.7} & 89.6 \gain{+2.3} & 28.8 \gain{+0.0} \\
\rowcolor{hyprow}
Qwen3-4B & Hypothesis & 93.4 \gain{+12.7} & \textbf{98.0} \gain{+16.0} & \textbf{100.0} \gain{+6.5} & 94.8 \gain{+5.1} & 87.2 \gain{+25.7} & 82.6 \gain{+8.7} & 49.3 \gain{+1.9} & 91.0 \gain{+3.7} & 29.9 \gain{+1.1} \\
\rowcolor{actrow}
Qwen3-4B & \method{} & \textbf{95.6} \gain{+14.9} & 97.0 \gain{+15.0} & \textbf{100.0} \gain{+6.5} & \textbf{98.3} \gain{+8.6} & \textbf{92.3} \gain{+30.8} & \textbf{89.1} \gain{+15.2} & \textbf{50.7} \gain{+3.3} & \textbf{92.5} \gain{+5.2} & \textbf{31.2} \gain{+2.4} \\
\bottomrule
\end{tabular}}
\caption{First-pass and final repair success rates in percent. Overall reports aggregate success; ALFWorld columns report task families and TextCraft columns report visible recipe depth. No skills and With skills are unrepaired first-pass rows; for TextCraft, With skills is the repair-pool R0 used for repair deltas. Direct, Hypothesis, and \method{} report final cumulative success after R7 for ALFWorld and R4 for TextCraft. Gray parentheses show change relative to the same model's With-skills/R0 row.}
\label{tab:main-results}
\end{table*}

Our experiments separate first-pass skill use, one-shot repair, and structured multi-round repair. All methods start from the same failure pool, edit the same skill-file format, and use frozen language models; improvements come only from external skill edits and retesting. We evaluate on two interactive benchmarks with complementary failure modes:
\begin{itemize}
\item ALFWorld contains 274 household tasks, split into 140 seen and 134 unseen tasks. For reporting, we group the two object-placement variants into a single Pick family and retain Look, Clean, Heat, and Cool as separate task families. The initial ALFWorld skill file is adapted from the ALFWorld skill substrate used by SKILL0~\citep{lu2026skill0}: each rollout receives general skills plus the mapped task-family section, and repair notes are appended to the corresponding local section. 
\item TextCraft uses the AgentGym TextCraft task set, where agents must follow visible recipe graphs and emit exact executable crafting actions. Because TextCraft has no external task-family skill file, we initialize it with a compact LLM-authored, recipe-free skill file containing only response-format, observation-grounding, exact-name, recipe-chain, and count-discipline constraints. We report TextCraft by visible recipe depth: D1 tasks require direct recipe use, while D2 tasks require one additional intermediate dependency. 
\end{itemize}

\paragraph{Models and scale.} Our primary evaluation uses Qwen3-1.7B, Qwen2.5-3B, and Qwen3-4B. Because repair is triggered only after an initial failure, informative evaluation requires a sufficiently large and diverse residual failure pool. As first-pass performance approaches ceiling, fewer tasks remain eligible for repair and the headroom for comparing repair methods decreases. We therefore focus on open-weight models in the 1.7B--4B range, which represent locally deployable agents while providing sufficient repair opportunities across capability levels.

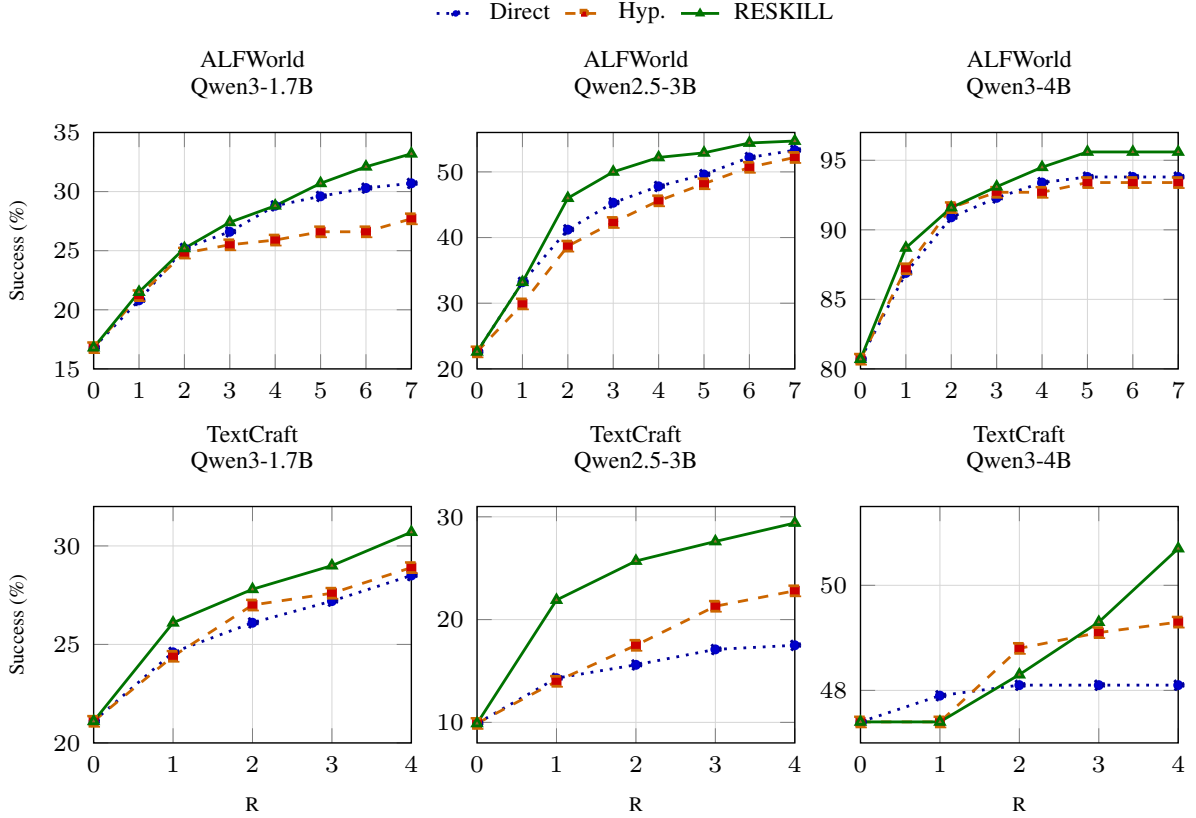
\begin{figure*}[t]
\centering
\resizebox{\textwidth}{!}{%
\begin{tikzpicture}
\begin{groupplot}[
  group style={
    group size=3 by 2,
    horizontal sep=0.042\textwidth,
    vertical sep=0.088\textwidth
  },
  width=0.302\textwidth,
  height=0.25\textwidth,
  grid=both,
  grid style={line width=.1pt, draw=gray!18},
  major grid style={line width=.25pt, draw=gray!32},
  tick label style={font=\scriptsize},
  label style={font=\tiny},
  title style={font=\scriptsize, align=center},
  legend style={
    font=\scriptsize\selectfont,
    draw=none,
    fill=none,
    inner sep=1pt,
    nodes={inner sep=1pt},
    column sep=3pt,
  },
  legend image post style={scale=0.65},
  legend columns=3,
  ylabel={Success (\%)},
  every axis plot/.append style={line width=0.82pt, mark size=1.45pt},
]

\nextgroupplot[
  title={\shortstack{ALFWorld\\Qwen3-1.7B}},
  xmin=0, xmax=7,
  ymin=15, ymax=35,
  xtick={0,1,2,3,4,5,6,7},
  legend to name=budgetlegend
]
\addplot+[blue!60!black, dotted, mark=*] coordinates {
  (0,16.8) (1,20.8) (2,25.2) (3,26.6) (4,28.8) (5,29.6) (6,30.3) (7,30.7)
};
\addlegendentry{Direct}

\addplot+[orange!80!black, dashed, mark=square*] coordinates {
  (0,16.8) (1,21.2) (2,24.8) (3,25.5) (4,25.9) (5,26.6) (6,26.6) (7,27.7)
};
\addlegendentry{Hyp.}

\addplot+[green!45!black, mark=triangle*] coordinates {
  (0,16.8) (1,21.5) (2,25.2) (3,27.4) (4,28.8) (5,30.7) (6,32.1) (7,33.2)
};
\addlegendentry{RESKILL}

\nextgroupplot[
  title={\shortstack{ALFWorld\\Qwen2.5-3B}},
  xmin=0, xmax=7,
  ymin=20, ymax=56,
  xtick={0,1,2,3,4,5,6,7},
  ylabel={}
]
\addplot+[blue!60!black, dotted, mark=*] coordinates {
  (0,22.6) (1,33.2) (2,41.2) (3,45.3) (4,47.8) (5,49.6) (6,52.2) (7,53.3)
};
\addplot+[orange!80!black, dashed, mark=square*] coordinates {
  (0,22.6) (1,29.9) (2,38.7) (3,42.3) (4,45.6) (5,48.2) (6,50.7) (7,52.2)
};
\addplot+[green!45!black, mark=triangle*] coordinates {
  (0,22.6) (1,33.2) (2,46.0) (3,50.0) (4,52.2) (5,52.9) (6,54.4) (7,54.7)
};

\nextgroupplot[
  title={\shortstack{ALFWorld\\Qwen3-4B}},
  xmin=0, xmax=7,
  ymin=80, ymax=97,
  xtick={0,1,2,3,4,5,6,7},
  ylabel={}
]
\addplot+[blue!60!black, dotted, mark=*] coordinates {
  (0,80.7) (1,86.9) (2,90.9) (3,92.3) (4,93.4) (5,93.8) (6,93.8) (7,93.8)
};
\addplot+[orange!80!black, dashed, mark=square*] coordinates {
  (0,80.7) (1,87.2) (2,91.6) (3,92.7) (4,92.7) (5,93.4) (6,93.4) (7,93.4)
};
\addplot+[green!45!black, mark=triangle*] coordinates {
  (0,80.7) (1,88.7) (2,91.6) (3,93.1) (4,94.5) (5,95.6) (6,95.6) (7,95.6)
};

\nextgroupplot[
  title={\shortstack{TextCraft\\Qwen3-1.7B}},
  xmin=0, xmax=4,
  ymin=20, ymax=32,
  xtick={0,1,2,3,4},
  xlabel={R}
]
\addplot+[blue!60!black, dotted, mark=*] coordinates {
  (0,21.1) (1,24.6) (2,26.1) (3,27.2) (4,28.5)
};
\addplot+[orange!80!black, dashed, mark=square*] coordinates {
  (0,21.1) (1,24.4) (2,27.0) (3,27.6) (4,28.9)
};
\addplot+[green!45!black, mark=triangle*] coordinates {
  (0,21.1) (1,26.1) (2,27.8) (3,29.0) (4,30.7)
};

\nextgroupplot[
  title={\shortstack{TextCraft\\Qwen2.5-3B}},
  xmin=0, xmax=4,
  ymin=8, ymax=31,
  xtick={0,1,2,3,4},
  xlabel={R},
  ylabel={}
]
\addplot+[blue!60!black, dotted, mark=*] coordinates {
  (0,9.9) (1,14.3) (2,15.6) (3,17.1) (4,17.5)
};
\addplot+[orange!80!black, dashed, mark=square*] coordinates {
  (0,9.9) (1,14.0) (2,17.5) (3,21.3) (4,22.8)
};
\addplot+[green!45!black, mark=triangle*] coordinates {
  (0,9.9) (1,21.9) (2,25.7) (3,27.6) (4,29.4)
};

\nextgroupplot[
  title={\shortstack{TextCraft\\Qwen3-4B}},
  xmin=0, xmax=4,
  ymin=47, ymax=51.5,
  xtick={0,1,2,3,4},
  xlabel={R},
  ylabel={}
]
\addplot+[blue!60!black, dotted, mark=*] coordinates {
  (0,47.4) (1,47.9) (2,48.1) (3,48.1) (4,48.1)
};
\addplot+[orange!80!black, dashed, mark=square*] coordinates {
  (0,47.4) (1,47.4) (2,48.8) (3,49.1) (4,49.3)
};
\addplot+[green!45!black, mark=triangle*] coordinates {
  (0,47.4) (1,47.4) (2,48.3) (3,49.3) (4,50.7)
};

\end{groupplot}

\node[anchor=south, yshift=9mm] at (group c2r1.north)
  {\pgfplotslegendfromname{budgetlegend}};

\end{tikzpicture}
}
\caption{Cumulative repair-budget trends. Each panel fixes one benchmark--model setting and plots success rate as the repair budget \(R\) increases. ALFWorld panels run through R7; TextCraft panels run through R4.}
\label{fig:budget-trends}
\end{figure*}

We compare \method{} with two single-patch repair procedures. Direct Repair generates a single local patch at each repair round from the current failed trajectory before retesting. Hypothesis Repair first generates failure hypotheses at each round, then writes one local patch conditioned on them. Neither baseline builds a multi-candidate repair pool, a coverage matrix, or an outcome-conditioned repair state. In contrast, \method{} keeps multiple hypotheses and candidate edits, selects by weighted coverage, and carries failed retest evidence into later rounds. ALFWorld is tracked through seven repair rounds, while TextCraft is tracked through four. In Table~\ref{tab:main-results}, No skills~\cite{li2026skillsbench} and With skills are unrepaired first-pass results; the three repair rows report final cumulative success after the full repair budget, i.e., R7 for ALFWorld and R4 for TextCraft. Figure~\ref{fig:budget-trends} and Appendix~\ref{app:budget-curves} show the intermediate budgets.

\subsection{Skill Use and Repair Effects}

\paragraph{Skills provide useful but incomplete guidance.}
Table~\ref{tab:main-results} first separates skill use from post-failure repair. In ALFWorld, the With skills row improves most task families for all three model sizes, with especially large gains on Look and Clean for the smaller models. TextCraft is more uneven: skills can improve depth-1 behavior, but the repair-pool R0 remains low for smaller models, and depth-2 tasks remain difficult. This pattern motivates post-failure editing: skills provide useful structure, but a failed rollout can still expose missing conditions, overly broad rules, or brittle surface-form behavior.

\paragraph{One-shot repair works when the trace exposes a single correction.}

Direct Repair is strongest when the failure feedback already points to a concrete local edit. In ALFWorld, it matches or exceeds \method{} on several family columns: it is strongest on Qwen3-1.7B Heat, Qwen2.5-3B Pick and Clean, and ties Qwen3-4B Heat. These cases often involve a specific procedural omission or action-interface mismatch, where a single patch can be sufficient. This pattern identifies a natural strength of one-shot repair: when the evidence localizes a single correction, direct editing can recover quickly. The complementary regime is where \method{} is most useful: the failed trace supports several plausible edits, or an unsuccessful retest should redirect the next repair rather than trigger another unconstrained rewrite.

\paragraph{Structured attribution helps distinguish between plausible repairs.}
\method{} obtains the best full-suite ALFWorld score for all three model sizes while also giving the best TextCraft D1 and D2 results for all three models. Its advantage is clearest when the repair choice is underdetermined. In ALFWorld, this appears in families where a failed trajectory can implicate object search, precondition handling, action ordering, or family scope. In TextCraft, surface form and planning interact: a wrong action may reveal an ingredient-string mismatch, but a useful skill edit may also need to preserve counts and follow the visible recipe chain. Direct Repair often captures one constraint, and Hypothesis Repair can name the failure more explicitly; \method{} compares candidate edits by the hypotheses they cover before committing to a patch.

\paragraph{Model strength changes the repair regime.}
The smaller models leave larger recoverable failure pools, so the benefit of repair is more visible across both domains. The largest TextCraft margin appears for Qwen2.5-3B, where \method{} improves both D1 and D2 substantially over the With skills row and over the two single-patch baselines. Qwen3-4B starts from stronger first-pass behavior, so the remaining repair pool is smaller and family-level differences are often close. Even in this higher-ceiling regime, \method{} remains best on TextCraft D1/D2 and reaches the highest ALFWorld full-suite repair result, with gains concentrated in harder families such as Clean and Cool.

\subsection{Repair-Budget Dynamics}

Figure~\ref{fig:budget-trends} plots cumulative success as the repair budget \(R\) increases. The curves reinforce the boundary suggested by Table~\ref{tab:main-results}. Direct Repair is often competitive in the first one or two rounds because many traces expose a concrete surface error that a single patch can fix. When that first patch is correct, the direct baseline recovers quickly. When it is incomplete, however, the direct and hypothesis-conditioned curves more often flatten, because the next prompt has little structured record of which explanation was tried and how the retest changed the evidence.

The trend is especially visible for TextCraft Qwen2.5-3B, where \method{} jumps sharply after the first repair round and keeps a large margin through R4. TextCraft failures often combine recipe-chain selection, exact surface-form copying, and count discipline, so recording which explanation a repair targeted makes a failed retest useful: the next round can emphasize remaining explanations instead of rewriting the same general rule. For Qwen3-4B TextCraft, stronger first-pass performance leaves a smaller residual failure pool and less room for improvement, resulting in closer repair curves. Even when there is limited room for improvement, \method{} achieves the highest final success, indicating that structured selection remains useful for difficult residual failures. 

ALFWorld shows the complementary pattern. Smaller models benefit steadily across later rounds, while Qwen3-4B approaches saturation by the middle of the budget. The important point is not only the final percentage, but the shape of the curve: \method{} is less dependent on the first patch being correct, because the post-repair trajectory can redirect later edits toward persistent or newly exposed failures. The budget curves also provide a practical reference for choosing skill-repair strategies under different repair-round budgets. Appendix~\ref{app:budget-curves} reports the full budget tables, separated by benchmark because ALFWorld is tracked through R7 whereas TextCraft is tracked through R4.

\subsection{Failure Modes and Repair Behavior}

The repair traces help explain why the same framework behaves differently across domains. Direct Repair and Hypothesis Repair both commit to a model-selected edit at each round. \method{} instead exposes the decision as a comparison among local repairs and carries retest feedback into the next state. This matters most when early diagnoses are noisy: for Qwen2.5-3B TextCraft, all methods begin from the same 9.9\% first-pass success rate, but \method{} reaches 21.9\% after one repair round, while Direct and Hypothesis Repair reach 14.3\% and 14.0\%. The gap persists through R4, suggesting that structured selection and outcome-conditioned update help the agent use post-failure feedback more effectively.

TextCraft depth stratification also clarifies where the gains come from. Most recoveries occur in depth-1 and depth-2 tasks, where the visible observation exposes a short recipe chain and small surface mistakes can invalidate an otherwise plausible plan. These tasks reward repairs that preserve exact item names, numeric counts, and command form. Appendix~\ref{app:case-study} gives a representative TextCraft trace: the initial agent rewrites a listed recipe token, while \method{} separates the surface-form hypothesis from broader planning hypotheses and selects a patch that directly targets the exposed mismatch.

For ALFWorld, trace inspection points to a different pattern. Direct patches can work well when the failure is a single procedural omission, such as not checking the right receptacle or not applying a family-specific action. \method{} is most useful when a failed household trajectory leaves several explanations plausible: object search, precondition handling, action ordering, or family scope. In those cases, the coverage relation gives the selector a reason to prefer a patch that addresses more of the active explanation mass, and an unsuccessful retest changes the next-round state rather than simply asking for another unconstrained rewrite.

\section{Conclusion}
\label{sec:conclusion}

We presented \method{}, a structured framework for post-failure skill repair. By tracking competing failure hypotheses, targeted patches, retest outcomes, and repair credit across rounds, \method{} turns failed repairs into feedback for later edits. Experiments on ALFWorld and TextCraft show gains over direct and hypothesis-conditioned repair across three model sizes, with the benefits when failures admit multiple plausible explanations.

\section{Acknowledgements}
This work is supported by Advanced Materials-National Science and Technology Major Project (Grant No. 2025ZD0620100), National Key R\&D Program of China (No. 2024YFA1012700), and Guangdong Provincial Key Lab of Integrated Communication, Sensing and Computation for Ubiquitous Internet of Things (No. 2023B1212010007).
\section{Limitations}
\label{sec:limitations}

Our evaluation studies interactive environments where failures produce observable action feedback and repaired skills can be retested under matched budgets. This scope supports controlled comparison of repair decisions and makes it possible to inspect how hypotheses, selected patches, and retest outcomes affect later rounds. Broader tool-use settings may introduce delayed feedback, noisier failure signals, or edits whose effects appear across longer interaction histories; extending structured repair to those settings is an important direction for future work.

\bibliography{custom}

\appendix
\section{Implementation Details}
\label{app:implementation}

This appendix documents the configuration of every reported experimental setting. Table~\ref{tab:provenance} maps each result to its task pool, model configuration, initial skill file, decoding parameters, context and action limits, transcript policy, repair budget, task-level logs, and reproduction command. 

\subsection{Generation Settings}

All reported experiments use frozen model weights served through an OpenAI-compatible local inference endpoint. Environment rollouts are decoded greedily. For ALFWorld first-pass rollouts and repair retests, we set the LLM-agent decoding temperature to \(0.0\) and top-\(p\) to \(0.7\), with a 512-token action-generation cap, seed 0, and an 8192-token serving context. ALFWorld repair-stage generations use stage-specific settings: failure hypotheses use temperature \(0.2\), top-\(p=0.95\); repair proposals use temperature \(0.6\), top-\(p=0.95\); and coverage construction uses temperature \(0.1\), top-\(p=0.9\). Each ALFWorld repair-stage call uses a 2048-token completion cap in the reported runs.

For TextCraft, environment rollouts also use greedy decoding with top-\(p=1.0\) and at most ten interaction turns. The full TextCraft setting uses a 4096-token serving context and a 128-token action-generation cap, while the Qwen3-4B depth-1/depth-2 setting uses an 8192-token serving context and a 256-token action-generation cap. TextCraft structured repair calls use top-\(p=1.0\) and a JSON completion cap of 768 tokens. Hypothesis, coverage, and post-repair attribution calls use temperature \(0.0\) under the compact JSON profile. Single-patch proposal calls use temperature \(0.2\) for Qwen3 models and \(0.3\) for Qwen2.5-3B; multi-proposal RESKILL calls use temperature \(0.2\) for Qwen3 models and \(0.5\) for Qwen2.5-3B. Failed JSON-format attempts are retried with temperature \(0.0\).

\subsection{ALFWorld Implementation}

The ALFWorld evaluation contains 274 tasks, including 140 seen and 134 unseen tasks. Within each model setting, Direct Repair, Hypothesis Repair, and \method{} share the same first-pass trajectories, initial skill file, evaluator, decoding configuration, and repair budget. The Qwen2.5-3B seen and unseen results are aggregated over the same task pools used by all three repair methods.

The shared repair loop builds failure packets, generates hypotheses and repairs, filters candidate edits, constructs coverage matrices, applies selected repair notes, retests failed tasks, runs post-repair attribution, and writes budget summaries. 


\paragraph{Skill editing.}
ALFWorld skills are stored as task-family sections. Each rollout receives general guidance plus the section mapped to the current family. The two picking variants share one picking section, while inspection, cleaning, heating, and cooling use separate sections. Repairs are appended as local notes rather than rewriting the entire skill file.


\subsection{TextCraft Implementation}

The Qwen3-1.7B and Qwen2.5-3B TextCraft experiments use the full task pool, recipe-free skills, a 4096-token context, and a 128-token action limit. The Qwen3-4B experiment uses the depth-1 and depth-2 task pool, an 8192-token context, and a 256-token action limit; its agent-visible trajectories contain only environment observations and actions. Within each model setting, Direct Repair, Hypothesis Repair, and \method{} share the same task pool, initial skill file, evaluator, decoding configuration, transcript policy, and repair budget. The Qwen3-4B results therefore support controlled comparisons among repair methods within that setting, but are not used for strict cross-model comparisons with the 1.7B and 3B results. The corresponding configurations, task-level logs, and reproduction commands are listed in Table~\ref{tab:provenance}.


\paragraph{Recipe-free skills.}
The TextCraft skill file contains only general response constraints: follow environment instructions, output one action line, include numeric quantities, use visible item names and craft commands, avoid external Minecraft knowledge, and react to execution failures. It contains no task-specific recipes.

\paragraph{Visible transcript policy.}
In the final 4B setting, any model-generated analysis or rationale preceding the executable action is treated as internal inference and excluded from the trajectory passed to subsequent model calls or the environment. The visible transcript therefore contains only environment observations and cleaned action text. Hypothesis generation, proposal generation, and coverage scoring condition on this cleaned transcript. Repair retesting and post-repair attribution may still use internal reasoning, but only cleaned actions and structured repair outputs are retained as interaction evidence.

\section{Repair Budget Tables and Trends}
\label{app:budget-curves}

R0 is the unrepaired with-skills first pass for each repair curve. For ALFWorld, this is the full-suite with-skills baseline in Table~\ref{tab:main-results}; for TextCraft, it is the recipe-free first pass in the corresponding full or depth-filtered repair pool. Table~\ref{tab:budget-alfworld} reports ALFWorld through R7, while Table~\ref{tab:budget-textcraft} reports TextCraft through R4.

\begin{table*}[t]
\centering
\scriptsize
\setlength{\tabcolsep}{3.2pt}
\renewcommand{\arraystretch}{1.12}
\resizebox{\textwidth}{!}{
\begin{tabular}{@{}llrrrrrrrr@{}}
\toprule
Model & Method & R0 & R1 & R2 & R3 & R4 & R5 & R6 & R7 \\
\midrule
Qwen3-1.7B & Direct & 16.8 & 20.8 \gain{+4.0} & 25.2 \gain{+8.4} & 26.6 \gain{+9.8} & 28.8 \gain{+12.0} & 29.6 \gain{+12.8} & 30.3 \gain{+13.5} & 30.7 \gain{+13.9} \\
Qwen3-1.7B & Hypothesis & 16.8 & 21.2 \gain{+4.4} & 24.8 \gain{+8.0} & 25.5 \gain{+8.7} & 25.9 \gain{+9.1} & 26.6 \gain{+9.8} & 26.6 \gain{+9.8} & 27.7 \gain{+10.9} \\
\rowcolor{actrow}
Qwen3-1.7B & \method{} & 16.8 & \textbf{21.5} \gain{+4.7} & \textbf{25.2} \gain{+8.4} & \textbf{27.4} \gain{+10.6} & \textbf{28.8} \gain{+12.0} & \textbf{30.7} \gain{+13.9} & \textbf{32.1} \gain{+15.3} & \textbf{33.2} \gain{+16.4} \\
\midrule
Qwen2.5-3B & Direct & 22.6 & 33.2 \gain{+10.6} & 41.2 \gain{+18.6} & 45.3 \gain{+22.7} & 47.8 \gain{+25.2} & 49.6 \gain{+27.0} & 52.2 \gain{+29.6} & 53.3 \gain{+30.7} \\
Qwen2.5-3B & Hypothesis & 22.6 & 29.9 \gain{+7.3} & 38.7 \gain{+16.1} & 42.3 \gain{+19.7} & 45.6 \gain{+23.0} & 48.2 \gain{+25.6} & 50.7 \gain{+28.1} & 52.2 \gain{+29.6} \\
\rowcolor{actrow}
Qwen2.5-3B & \method{} & 22.6 & \textbf{33.2} \gain{+10.6} & \textbf{46.0} \gain{+23.4} & \textbf{50.0} \gain{+27.4} & \textbf{52.2} \gain{+29.6} & \textbf{52.9} \gain{+30.3} & \textbf{54.4} \gain{+31.8} & \textbf{54.7} \gain{+32.1} \\
\midrule
Qwen3-4B & Direct & 80.7 & 86.9 \gain{+6.2} & 90.9 \gain{+10.2} & 92.3 \gain{+11.6} & 93.4 \gain{+12.7} & 93.8 \gain{+13.1} & 93.8 \gain{+13.1} & 93.8 \gain{+13.1} \\
Qwen3-4B & Hypothesis & 80.7 & 87.2 \gain{+6.5} & \textbf{91.6} \gain{+10.9} & 92.7 \gain{+12.0} & 92.7 \gain{+12.0} & 93.4 \gain{+12.7} & 93.4 \gain{+12.7} & 93.4 \gain{+12.7} \\
\rowcolor{actrow}
Qwen3-4B & \method{} & 80.7 & \textbf{88.7} \gain{+8.0} & \textbf{91.6} \gain{+10.9} & \textbf{93.1} \gain{+12.4} & \textbf{94.5} \gain{+13.8} & \textbf{95.6} \gain{+14.9} & \textbf{95.6} \gain{+14.9} & \textbf{95.6} \gain{+14.9} \\
\bottomrule
\end{tabular}}
\caption{ALFWorld cumulative repair-budget success rates in percent. Parentheses show the gain relative to R0 within the same row; shaded rows mark \method{}.}
\label{tab:budget-alfworld}
\end{table*}

\begin{table*}[t]
\centering
\normalsize
\setlength{\tabcolsep}{8pt}
\renewcommand{\arraystretch}{1.16}
\resizebox{0.98\textwidth}{!}{
\begin{tabular}{@{}llrrrrr@{}}
\toprule
Model & Method & R0 & R1 & R2 & R3 & R4 \\
\midrule
Qwen3-1.7B & Direct & 21.1 & 24.6 \gain{+3.5} & 26.1 \gain{+5.0} & 27.2 \gain{+6.1} & 28.5 \gain{+7.4} \\
Qwen3-1.7B & Hypothesis & 21.1 & 24.4 \gain{+3.3} & 27.0 \gain{+5.9} & 27.6 \gain{+6.5} & 28.9 \gain{+7.8} \\
\rowcolor{actrow}
Qwen3-1.7B & \method{} & 21.1 & \textbf{26.1} \gain{+5.0} & \textbf{27.8} \gain{+6.7} & \textbf{29.0} \gain{+7.9} & \textbf{30.7} \gain{+9.6} \\
\midrule
Qwen2.5-3B & Direct & 9.9 & 14.3 \gain{+4.4} & 15.6 \gain{+5.7} & 17.1 \gain{+7.2} & 17.5 \gain{+7.6} \\
Qwen2.5-3B & Hypothesis & 9.9 & 14.0 \gain{+4.1} & 17.5 \gain{+7.6} & 21.3 \gain{+11.4} & 22.8 \gain{+12.9} \\
\rowcolor{actrow}
Qwen2.5-3B & \method{} & 9.9 & \textbf{21.9} \gain{+12.0} & \textbf{25.7} \gain{+15.8} & \textbf{27.6} \gain{+17.7} & \textbf{29.4} \gain{+19.5} \\
\midrule
Qwen3-4B & Direct & 47.4 & \textbf{47.9} \gain{+0.5} & 48.1 \gain{+0.7} & 48.1 \gain{+0.7} & 48.1 \gain{+0.7} \\
Qwen3-4B & Hypothesis & 47.4 & 47.4 \gain{+0.0} & \textbf{48.8} \gain{+1.4} & 49.1 \gain{+1.7} & 49.3 \gain{+1.9} \\
\rowcolor{actrow}
Qwen3-4B & \method{} & 47.4 & 47.4 \gain{+0.0} & 48.3 \gain{+0.9} & \textbf{49.3} \gain{+1.9} & \textbf{50.7} \gain{+3.3} \\
\bottomrule
\end{tabular}}
\caption{TextCraft cumulative repair-budget success rates in percent. Parentheses show the gain relative to R0 within the same row; shaded rows mark \method{}. TextCraft is tracked through R4, so this table intentionally omits later repair budgets.}
\label{tab:budget-textcraft}
\end{table*}

\section{Evaluation Setting Summary}

\begin{table*}[t]
\centering
\small
\renewcommand{\arraystretch}{1.12}
\begin{tabular}{p{0.24\linewidth}p{0.68\linewidth}}
\toprule
Reported setting & Evaluation configuration \\
\midrule
ALFWorld Qwen3-1.7B and Qwen3-4B &
274 tasks, including 140 seen and 134 unseen tasks; 8192-token context; 512-token action limit; repair budget \(R=7\). Repair methods share the same first-pass trajectories and initial skill file within each model setting. \\
ALFWorld Qwen2.5-3B &
The same 140 seen and 134 unseen tasks; 8192-token context; 512-token action limit; repair budget \(R=7\). Repair methods share the same first-pass trajectories within each split. \\

TextCraft Qwen3-1.7B and Qwen2.5-3B &
Full TextCraft task pool with recipe-free skills, which contain general rules for reading recipes and executing actions but no task-specific crafting recipes; 4096-token context; 128-token action limit; at most ten interaction turns; repair budget \(R=4\). \\

TextCraft Qwen3-4B &
Depth-1 and depth-2 task pool with the same recipe-free skill design; 8192-token context; 256-token action limit; at most ten interaction turns; agent-visible trajectories containing only environment observations and actions; repair budget \(R=4\). \\

ALFWorld skills ablation &
Matched first-pass evaluations with and without the initial skill file over the same 140 seen and 134 unseen tasks for each model. \\

TextCraft skills ablation &
Matched first-pass evaluations with and without the recipe-free initial skill file over the same depth-1 and depth-2 task order for each model setting. \\

\bottomrule
\end{tabular}
\caption{
Evaluation configurations for the reported settings.
All repair methods are compared under matched conditions within each model and benchmark setting.
}
\label{tab:provenance}
\end{table*}

\section{Additional Notes on TextCraft Depth}

The TextCraft depth diagnostic is computed from visible recipe lines in the initial observation. The full-setting analysis includes depths one through four for the 1.7B and 3B settings. The final 4B evaluation focuses on the depth-1 and depth-2 subset and uses the same visible-recipe rule.

\section{Case Study: Auditable TextCraft Repair}
\label{app:case-study}

Table~\ref{tab:textcraft-case} shows a representative Qwen3-4B TextCraft example from the reported depth-1 and depth-2 setting. The goal is to craft spruce planks. Direct Repair and Hypothesis Repair remain unsuccessful after the repair budget on this item, while \method{} succeeds after one repair round.

\begin{table*}[t]
\centering
\small
\setlength{\tabcolsep}{5pt}
\renewcommand{\arraystretch}{1.12}
\begin{tabular}{p{0.17\linewidth}p{0.75\linewidth}}
\toprule
Field & Trace summary \\
\midrule
Task & Goal: spruce planks; visible recipe: ``craft 4 spruce planks using 1 spruce logs''; recipe depth: 1. \\
Trajectory excerpt & The observation lists the recipe as ``craft 4 spruce planks using 1 spruce logs.'' The failed action rewrites the ingredient as ``craft 4 spruce planks using 1 spruce log.'' The environment rejects the action as an invalid recipe. The important evidence is not that the agent chose the wrong goal, but that it changed the visible surface form of the ingredient. \\
Initial hypotheses & The first hypothesis states that the agent singularized a listed ingredient token instead of copying the recipe surface form. Other hypotheses cover generic recipe-format mismatch, direct-goal execution without checking prerequisites, and count discipline. The first-round weights are uniform because no repair has been tested yet. \\
Candidate repairs & One repair targets exact surface-form copying. Another targets visible recipe-chain execution. A third focuses on count preservation. These repairs overlap with different hypotheses, so the coverage matrix distinguishes a patch that fixes the exposed token mismatch from patches that only address broader recipe planning. \\
Selected repair & The selected patch tells the agent to treat listed TextCraft commands as exact surface forms: fetch missing ingredients using the exact listed item text and count, and craft by copying the visible recipe command rather than singularizing or pluralizing item names. \\
Retest outcome & The repaired skill snapshot succeeds on the retest after one repair round. The trace records both the chosen patch and the competing candidates, so the final success can be inspected as a consequence of the surface-form hypothesis and its selected repair. \\
\bottomrule
\end{tabular}
\caption{A TextCraft case study illustrating the audit trail produced by \method{}. The example is drawn from the final Qwen3-4B depth-1 and depth-2 run. It exposes the surface-form hypothesis, competing coverage-based repair choices, and the selected successful patch.}
\label{tab:textcraft-case}
\end{table*}

This case illustrates why the repair trace is useful beyond the final success flag. A direct patch can state a generic formatting rule without resolving the exact mismatch between ``spruce log'' and ``spruce logs.'' \method{} records the more specific surface-form hypothesis, compares candidate repairs through a binary coverage matrix, and preserves the retest outcome for the selected patch.

\clearpage
\section{Prompt Templates}
\label{app:prompts}

\newenvironment{prompttemplate}[1]{%
  \begin{tcolorbox}[
    colback=gray!20,
    colframe=gray!55,
    coltitle=black,
    colbacktitle=gray!35,
    boxrule=0.5pt,
    arc=2pt,
    left=5pt,
    right=5pt,
    top=5pt,
    bottom=5pt,
    title={#1},
    fonttitle=\bfseries\small,
    fontupper=\footnotesize
  ]%
}{%
  \end{tcolorbox}%
}

This appendix specifies the model inputs, instructions, and structured outputs
used at each stage of the repair process.

\subsection{Shared Evidence Bundle}

All repair-stage calls are grounded in a common set of observable task evidence, supplemented with stage-specific inputs such as active hypotheses, candidate patches, or the selected repair. The shared evidence includes:
\begin{itemize}
    \item the task goal and the initial environment observation visible to the agent;
    \item the relevant portion of the current skill set;
    \item a compact trajectory excerpt containing the agent's actions and the
    corresponding environment responses;
    \item benchmark-specific interface information available to the agent,
    such as visible recipes in TextCraft or valid action syntax in ALFWorld.
\end{itemize}
Only observable interaction content is treated as trajectory evidence. The prompts require each hypothesis and candidate patch to be grounded in this evidence, and require patches to express reusable local skill edits rather than episode-specific action sequences.

\begin{table*}[t]
\centering
\small
\setlength{\tabcolsep}{5pt}
\renewcommand{\arraystretch}{1.08}
\begin{tabular}{@{}p{0.12\textwidth}p{0.24\textwidth}p{0.56\textwidth}@{}}
\toprule
Symbol & Object & Role in \method{} \\
\midrule
\(S_t\) & Current skill set & Provides the editable skill context; selected repairs append local skill notes. \\
\(\tau_i^t\) & Failed or retest trajectory & Provides evidence for hypotheses, repair hints, and post-repair attribution. \\
\(\mathcal{H}_t\) & Hypothesis list & Finite support over candidate failure explanations. \\
\(b_t\) & State weights & Maintained by the algorithm over the current explanations. \\
\(\mathcal{Q}_t\) & Repair candidates & The finite set of reusable repairs considered at round \(t\). \\
\(C_t(q,h)\) & Coverage relation & Binary relation indicating whether repair \(q\) addresses hypothesized failure explanation \(h\). \\
\(q_t^\star\) & Selected repair & Chosen by weighted coverage plus fixed locality and grounding preferences. \\
\(b_{t+1}\) & Updated state & Computed from post-repair explanations and carried into the next repair round. \\
\bottomrule
\end{tabular}
\caption{Method variables used in the structured repair loop.}
\label{tab:prompt-method-map}
\end{table*}

\subsection{Repair Prompt Contracts}

\begin{prompttemplate}{Failure Attribution Prompt Template}
\textbf{Role}\newline
You are an analyst identifying concrete reasons why an interactive rollout failed.\newline
\textbf{Input Evidence}: \{task goal\}, \{current skill excerpt\}, \{failed trajectory\}, \{environment feedback\}.\newline
\textbf{Instructions}: List a small set of distinct failure explanations. Each explanation must cite observable evidence from the trajectory and indicate the skill scope it may affect. Avoid vague summaries and avoid proposing a repair in this step.\newline
\textbf{Expected Output}: Hypotheses with stable identifiers, evidence, affected scope, and repair direction.
\end{prompttemplate}

\begin{prompttemplate}{Direct Repair Prompt Template}
\textbf{Role}\newline
You are editing a reusable skill file after a failed rollout.\newline
\textbf{Input Evidence}: \{task goal\}, \{current skill excerpt\}, \{failed trajectory\}, \{environment feedback\}.\newline
\textbf{Instructions}: Write one local skill patch that could prevent the same type of failure while remaining compatible with the environment interface. Prefer reusable rules over one-episode action plans.\newline
\textbf{Expected Output}: One concise skill edit with target scope and appended rule.
\end{prompttemplate}

\begin{prompttemplate}{Hypothesis Repair Prompt Template}
\textbf{Role}\newline
You are editing a reusable skill file using an explicit failure explanation.\newline
\textbf{Input Evidence}: \{shared evidence bundle\}, \{current hypotheses \(\mathcal{H}_t\)\}.\newline
\textbf{Instructions}: Choose one useful hypothesis from the list and write a local patch that targets it. Keep the edit small, grounded, and compatible with the observed interface.\newline
\textbf{Expected Output}: Selected hypothesis identifier and one reusable skill edit.
\end{prompttemplate}

\begin{prompttemplate}{Post-Repair Attribution Prompt Template}
\textbf{Role}\newline
You are updating the repair state after a selected patch was retested and the task still failed.\newline
\textbf{Input Evidence}: \{previous hypotheses\}, \{selected repair \(q_t^\star\)\}, \{failed retest trajectory\}.\newline
\textbf{Instructions}: Explain whether the repair missed the issue, partially helped, or exposed a new failure. Reuse stable identifiers when possible and introduce a new hypothesis only when supported by the retest trace.\newline
\textbf{Expected Output}: Next-round hypotheses \(\mathcal{H}_{t+1}\) and a concise state-update rationale.
\end{prompttemplate}



\section{Information About Use Of AI Assistants}
This manuscript uses Ai Assistants strictly for the purpose of language editing and textual polishing to enhance presentation quality. We declare that the novel ideas, methodological framework, experimental execution, and data analysis are the original work of the authors. All content modified by AI tools has been carefully reviewed and validated by the authors to ensure accuracy.

\end{document}